\pdfoutput=1

\documentclass[11pt]{article}

\usepackage[final]{acl}

\usepackage{times}
\usepackage{latexsym}
\usepackage{amsmath}
\usepackage{multirow,makecell}
\usepackage{subfig}
\usepackage{hyperref}

\usepackage[T1]{fontenc}

\usepackage[utf8]{inputenc}

\usepackage{microtype}

\usepackage{inconsolata}

\usepackage{graphicx}

\title{NYK-MS: A Well-annotated Multi-modal Metaphor and Sarcasm Understanding Benchmark on Cartoon-Caption Dataset}

\author{Ke Chang$^{1, 2}$, Hao Li$^{1, 3}$, Junzhao Zhang$^{1, 3}$ \and Yunfang Wu$^{1, 2\dagger}$ \\
         $^1$National Key Laboratory for Multimedia Information Processing, Peking University\\
         $^2$School of Computer Science, Peking University\\
         $^3$School of Software and Microelectronics, Peking University \\
  \texttt{\{changkegg, fst1, wuyf\}@pku.edu.cn, 2301210532@stu.pku.edu.cn} \\
}

\begin{document}
\maketitle
\begin{abstract}
Metaphor and sarcasm are common figurative expressions in people's communication, especially on the Internet or the memes popular among teenagers. We create a new benchmark named NYK-MS (NewYorKer for Metaphor and Sarcasm), which contains 1,583 samples for metaphor understanding tasks and 1,578 samples for sarcasm understanding tasks. These tasks include whether it contains metaphor/sarcasm, which word or object contains metaphor/sarcasm, what does it satirize and why does it contains metaphor/sarcasm, all of the 7 tasks are well-annotated by at least 3 annotators. We annotate the dataset for several rounds to improve the consistency and quality, and use GUI and GPT-4V to raise our efficiency. Based on the benchmark, we conduct plenty of experiments. In the zero-shot experiments, we show that Large Language Models (LLM) and Large Multi-modal Models (LMM) can't do classification task well, and as the scale increases, the performance on other 5 tasks improves. In the experiments on traditional pre-train models, we show the enhancement with augment and alignment methods, which prove our benchmark is consistent with previous dataset and requires the model to understand both of the two modalities. 
\end{abstract}

\section{Introduction}

People often express their idea with metaphor or sarcasm. For example, they may use aliases of some important people instead of their real name, which contain metaphor and can escape from being blocked by many social networks. Besides, when they want to show their strong negative sentiment on some events, they can use words contains positive sentiment and satirize it. On the Internet, these phenomenon is becoming more and more common, bringing huge challenge for models to understand people's real meaning.

Formally, metaphor means that a word or phrase's real meaning isn't same with its basic meaning \citep{lakoff2008metaphors,lagerwerf2008openness}. In linguistics studies, researchers use MIP \citep{group2007mip,steen2010method} and SPV \citep{wilks1975preferential,wilks1978making} theories to define metaphor, which also provide strategies for auto detection \citep{choi-etal-2021-melbert}. Similarly, sarcasm can be defined as that a word or phrase's real sentiment doesn't consist with its basic sentiment \citep{zhang-etal-2024-multi-modal}, and usually means using positive words to express negative sentiment. In multi-modal situations, the word and phrase could also be an object in the image.

To deal with the task of metaphor and sarcasm understanding, a well-annotated dataset is quite important. However, previous datasets have their weakness, such as annotated by mechanical rules \citep{cai-etal-2019-multi}, mainly focusing on texts or only using memes \citep{zhang-etal-2021-multimet}. In this paper, we create a new benchmark called NYK-MS, which can benefit research on these tasks. Our work can be described through the questions and answers below:
\\

\textbf{Q1: Why do we use cartoon-caption as origin dataset?}

In our early work, we tried to collect data from Twitter, but the Tweets don't meet our requirement. First, the ratio that a Tweet contains metaphor and sarcasm is very low (see Table \ref{tab:early}). There is less than $20\%$ samples contains metaphor, even if we use \texttt{\#metaphor} as crawling tag. Besides, using selected MVSA dataset \citep{niu2016sentiment} is a worse choice, where only about 13\% samples contains metaphor. Detailed crawling and selection method can be seen in Appendix \ref{sec:a1}.

\begin{figure}[ht]
\centering
  \includegraphics[width=0.6\linewidth]{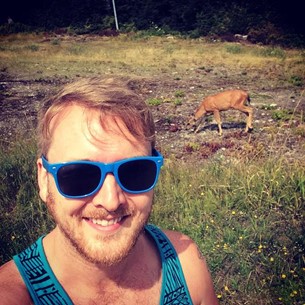} 
  \caption {An example in early annotation that contains metaphor. The text is "Photo {\color{red}bombed} by a deer, what an asshole."}
  \label{fig:twi_sample}
\end{figure}

Second, the metaphor and sarcasm in normal Tweets are quite easy, and the image is not necessary. Figure \ref{fig:twi_sample} shows a metaphor example. In this example, even if there is no photo, we can easily judge that the word "bombed" contains metaphor. Most of the positive examples in early annotation only contains single-modal metaphor or sarcasm, so they are not suitable with the multi-modal task.
\begin{table}[ht]
   \centering
   \small
    \renewcommand\arraystretch{1.2}
       
       \begin{tabular}{@{}ccccc@{}}
       \Xhline{1pt}
        \textbf{Data Source} & \textbf{Pos.} & \textbf{Neg.} & \textbf{Total} & \textbf{Pos. Rate}  \\ 
        \Xhline{0.5pt}
        Twitter  & 93 & 376 & 469 & 19.83\%\\
		MVSA  & 522 & 3476 & 4000 & 13.05\%\\
        \Xhline{0.5pt}
        All  & 615 & 3854 & 4469 & 13.76\%\\
        \Xhline{1pt}
       \end{tabular}
\caption{Early annotate results. Twitter means that the data is crawled from Twitter, and MVSA means that the data is selected from MVSA dataset \citep{niu2016sentiment}.}
\label{tab:early}
\end{table}

Third, limited by crawling settings, the crawled data always focus on specific events. For example, in MVSA dataset, a plenty of samples are related with \texttt{\#NationalDogDay} tag, so there are many positive metaphor words about dogs, such as "baby", "friend", "lover". However, our work doesn't focus on these topics, and we want to create a general dataset.

With these weakness, we finally give up using Tweets or MVSA dataset for annotation. Instead of these real corpus, \texttt{newyorker\_caption\_contest} \citep{hessel-etal-2023-androids} is a new dataset contains cartoon-caption pairs. The dataset is derived from \citet{newyorkernextmldataset}, \citet{shahaf2015inside}, and \citet{radev-etal-2016-humor}. It consists of 3 tasks: \textbf{Matching}, \textbf{Ranking} and \textbf{Explanation}. Since the cartoons are created by professional painters and always contain deep meaning, the ratio that they contains metaphor and sarcasm is far higher than the data shown above. Besides, the captions are selected from readers' submission, so the winners are always humor and concise, which requires the model must combine the two modals to understand the whole meaning, which is more difficult than Tweets. Finally, the cartoons are published in past several years, so they don't focus on specific topics. With these advantages, we choose this dataset as our origin data.
\\

\textbf{Q2: What tasks does NYK-MS support?}

Our NYK-MS dataset can be separated as NYK-M and NYK-S, supporting metaphor and sarcasm task respectively.

For metaphor understanding, NYK-M contains 3 tasks: 
\begin{itemize}
    \item Metaphor Classification (\textbf{MC}). Models should output 1 or 0, representing whether the sample contains metaphor.
    \item Metaphor Word detection (\textbf{MW}). Models should output the word (or phrase, object in image) that contains metaphor.
    \item Metaphor Explanation (\textbf{ME}). Models should output the explanation of the metaphor, including the literal meaning, the real meaning and the reason of such usage.
\end{itemize}

For sarcasm understanding, NYK-S contains 4 tasks. Besides the similar task \textbf{SC}  (Sarcasm Classification), \textbf{SW} (Sarcasm Word detection), \textbf{SE} (Sarcasm Explanation), we designed another task called \textbf{ST} (Sarcasm Target detection), which requires models to output the target of sarcasm, such as social phenomena or people. This task is important for application, because it can help models understand the position of speakers.

In conclusion, NYK-MS contains 7 tasks: MC, MW, ME, SC, SW, ST, SE. The detailed annotation workflow for these tasks is described in Section \ref{sec:workflow}.
\\





\textbf{Q3: What experiments do we conduct?}

For zero-shot situation, we use several LLMs and LMMs to do these task, and analyze their performance; For fine-tuning situation, we designed a baseline model following \citet{zhang-etal-2024-multi-modal} for MC and SC task, and using several alignment methods to improve the classification rate, including contrastive learning, Optimal Transport \citep{villani2009optimal}. Besides, we use data augmentation and knowledge augmentation method and get improvement, showing that our annotation standard is consistent with previous work and human understanding.
\\

Our contributions are as follow:

\begin{itemize}
    \item We create a new well-annotated benchmark NYK-MS for multi-modal metaphor and sarcasm understanding, including 7 tasks on more than 1,500 cartoon-caption pairs.
    \item We design a workflow to deal with the inconsistency of annotators, and use LMM's output as base annotation, changing the free writing task into checking and modifying task. This workflow can be used in any difficult task in which people's idea is not same.
    \item We conduct plenty of experiments, showing the performance of models, and use several methods to improve the performance on our NYK-MS dataset, including alignment and augmentation.
\end{itemize}

\section{Related Work}

\subsection{Previous Datasets}

For text-only metaphor understanding task, VUA18 \citep{leong2018report}, VUA20 \citep{leong2020report}, MOH-X \citep{mohammad2016metaphor}, TroFi \citep{birke2006clustering} are commonly used datasets. VUA series are annotated on  word level, tagging whether each word contains metaphor. For cross-modal task, MVSA \citep{niu2016sentiment} is a sentiment analysis dataset, setting -1, 0, 1 three sentiment and consists of MVSA\_Single and MVSA\_Multiple, the latter remains three origin annotation result. HFM \citep{cai-etal-2019-multi} is a large sarcasm detection dataset, but the train set is annotated by the tag \texttt{\#sarcasm}, which means if the text contains the tag, the result is 1, otherwise is 0. MET-Meme \citep{xu2022met} is a detailed dataset on meme, contains plenty of annotation items. MultiMET \citep{zhang-etal-2021-multimet} and MetaCLUE \citep{akula2023metaclue} are also multi-modal metaphor understanding datasets.

\subsection{Contrastive Learning}

SimCLR \citep{chen2020simple} and MoCo series \citep{he2020momentum} use contrastive in CV, and SimCSE uses it in NLP. CLIP \citep{radford2021learning} does alignment with contrastive learning, and ITC (Image-Text Contrastive) loss is used widely in recent multi-modal pre-training models, including BLIP \citep{li2022blip, li2023blip}.

\subsection{Optimal Transport}

OT (Optimal Transport) is a method to calculate a best transfer matrix between two distributions \citep{villani2009optimal}. \citet{cuturi2013sinkhorn} use an iteration method \citep{knight2008sinkhorn} to solve it effectively. Since the cross-modal alignment can be seen as a transfer process, Solving OT to calculate the transfer cost as loss function is used by some works \citep{pramanick2022multimodal, xu2023multimodal, aslam2024distilling}.

\section{Data Annotation Workflow}
\label{sec:workflow}
\subsection{Data Pre-process}
We download the \texttt{newyorker\_caption\_contest} dataset \citep{hessel-etal-2023-androids} from GitHub\footnote{See \href{https://github.com/jmhessel/caption_contest_corpus}{https://github.com/jmhessel/caption\_contest\_corpus}.}. As mentioned above, the dataset contains 3 tasks:
\begin{itemize}
    \item \textbf{Matching}. Giving a cartoon and 5 captions of different cartoons, models should choose the only related caption. This task is a 5-choice question, so only the right answer is useful for our annotation.
    \item \textbf{Ranking}. Giving a cartoon and 2 captions of it, models should choose the better caption. In this task, both of the 2 answers can be used for our annotation.
    \item \textbf{Explanation}. Giving a cartoon and its caption, models should output the explanation of them. The task can be used for our annotation, but the number of it is quite small.
\end{itemize}

We put all the images into a folder, and then analyze the JSON files. Table \ref{tab:analysis} shows the result of the 3 files for each task after data retrieval and deduplication:

\begin{table}[ht]
   \centering
    \renewcommand\arraystretch{1.2}

       \begin{small}
       \begin{tabular}{@{}cccc@{}}
       \Xhline{1pt}
        \textbf{Task} & \textbf{Cartoons} & \textbf{Captions} & \textbf{Cap. / Car.}  \\ 
        \Xhline{0.5pt}
        \textbf{Matching}  & 704 & 2653 & 3.77 (Average)\\
		\textbf{Ranking}  & 679 & 5127 & 7.54 (Average)\\
		\textbf{Explanation}  & 651 & 651 & 1\\
        \Xhline{1pt}
       \end{tabular}
       \end{small}
\caption{Analysis result for origin data. We use only the correct answer in Matching task, and the two answers in Ranking task. Cap. / Car. means the number of captions for each cartoon, and in Matching and Ranking task, the number is not fixed.}
\label{tab:analysis}
\end{table}

Besides, we get the union of 3 tasks, and the full set contains 704 cartoons and 5200 captions. We can see that the Ranking task contains 679 cartoons and 5127 captions, almost covering the full set, so we use the data in Ranking task for annotation. We use the 679 cartoons, and selected 3 captions randomly for each cartoon, so the origin dataset contains $679 \times 3=2037$ samples. For each sample, we use the cartoon, the caption and the description from the origin dataset for next step.

\subsection{Annotation GUI}

We designed two versions of  GUI using Tkinter library in Python. 
The first version is used in classification step, and the second version is used in modifying step. Fig \ref{fig:gui2} shows the GUI, and the details about them are shown in Appendix \ref{sec:a2}.

\begin{figure*}[ht]
\centering
  \includegraphics[width=0.8\linewidth]{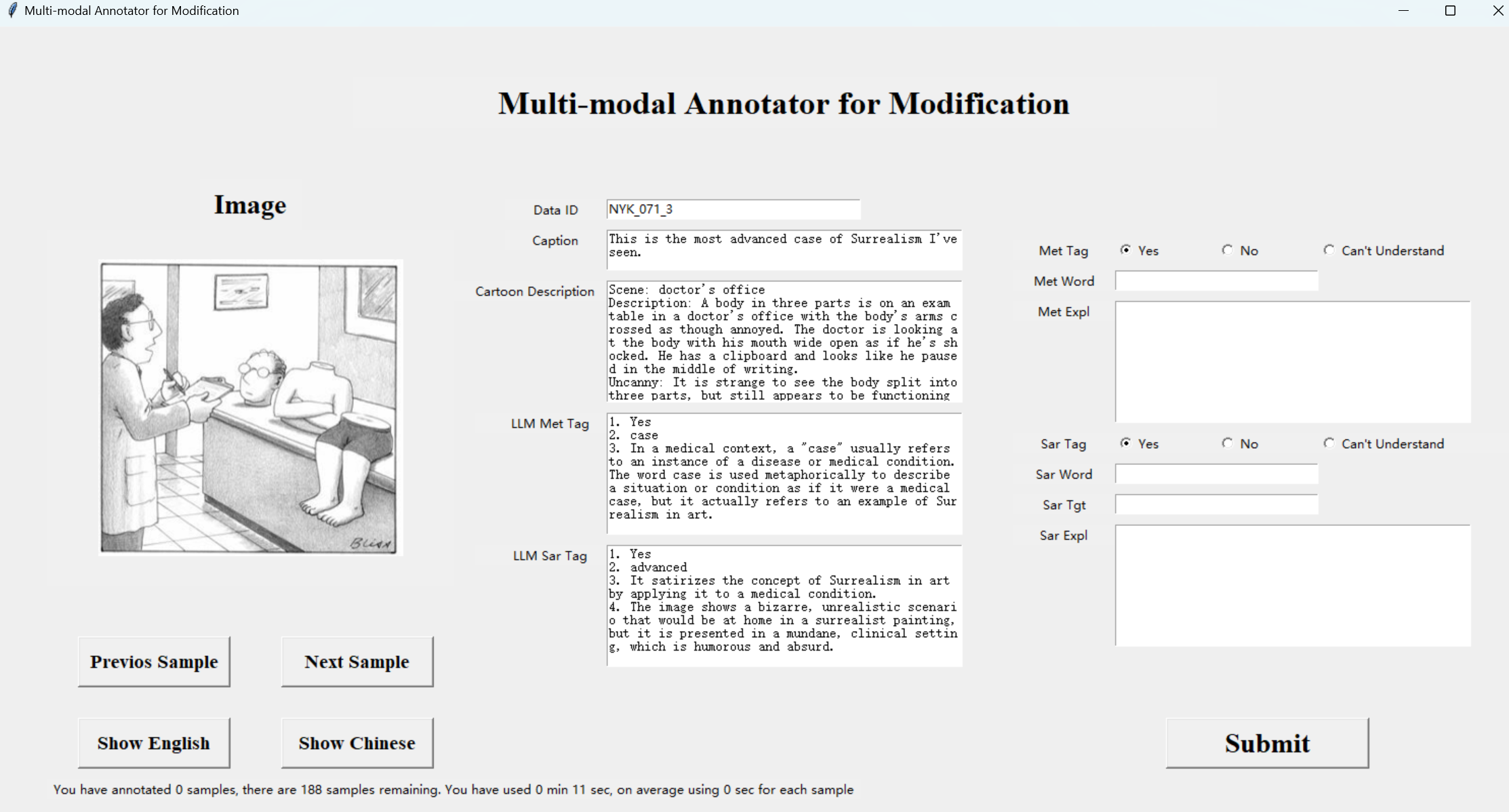}

  \caption {GUI for modifying step.}
    \label{fig:gui2}
\end{figure*}

\subsection{Classification Step}

With the processed data and GUI, we hired 12 annotators to do this step. These annotators are separated into 4 group, each group annotates $679 / 4 = 170$ (the last group is 169) cartoons. In this step, we regard the cartoon and the 3 captions as a whole, and only require the annotators to judge whether them contains deep meaning, no matter it is metaphor or sarcasm. However, after the annotation, we find that the consistency is very low. Table \ref{tab:consist} shows the result, and the consistency in round 1 is unacceptable. Almost all pair-wise Kappa is less than 0.2, meaning no consistency at all in statistic.

So, we discuss online to deal with the consistency problem. We selected 2 inconsistency (001 or 011) samples from each group's annotation result, and invite the annotator who think it is positive to explain his opinion. More details about this discussion can be seen in Appendix \ref{sec:a3}.

After the discussion, annotators start the second round annotation. For the 000 and 111 samples, we use the result from round 1, and only annotate 001 and 011 samples, which contains of $184+241=425$ samples. After this round, the consistency becomes higher, and we use the 011 and 111 samples ($123+259=382$ cartoons, $382 \times 3=1146$ cartoon-caption pairs) as positive samples for next step.
\begin{table*}[ht]
   \centering
   \small
    \renewcommand\arraystretch{1.2}
       
       \begin{tabular}{@{}c|c|c|cccc@{}}
       \Xhline{1pt}
        \textbf{Round} & \textbf{Group} & \textbf{Pair-wise Kappa} & \textbf{000} & \textbf{001} & \textbf{011} & \textbf{111}  \\ 
        \Xhline{0.5pt}
        \multirow{5}{*}{\textbf{Round 1}} & Group 1 & 0.0671, 0.0592, 0.2095&		7&		45&		62&		56\\
         & Group 2 & 	0.1164, 0.0608, 0.4068&	25&	62&	46&	37\\
          & Group 3 &0.1603, -0.0583, 0.0172&	11&	48&	70&	40\\
           & Group 4 & 	0.0730, 0.1732, 0.2929&	12&	29&	63&	66\\
            \cline{2-7}
           & Total & - & 55&	184&	241	&199\\
        \Xhline{0.5pt}
		\multirow{5}{*}{\textbf{Round 2}} & Group 1 & 0.3146, 0.2809, 0.4758&	21&	39&	36&	74\\
         & Group 2 & 	0.3360, 0.3744, 0.4826&	37&	45&	32&	56\\
          & Group 3 &0.4821, 0.4414, 0.5020&	54&	41&	26&	48\\
           & Group 4 &0.4226, 0.4226, 0.5713&	28&	32&	29&	81\\
            \cline{2-7}
           & Total & - & 140&	157&	123	&259\\
        \Xhline{1pt}
       \end{tabular}
\caption{Consistency in classification step. The pair-wise kappa means the Kappa between annotator 1 and 2, 1 and 3, 2 and 3. The 000, 001, 011, 111 means the distribution of annotation results, for example, 001 means 1 annotator thinks it is positive, and another 2 annotators think it is negative.}
\label{tab:consist}
\end{table*}

\subsection{GPT-4V Annotation Step}

With the 1146 positive samples, we use GPT-4V (through the API provided by Close-AI\footnote{See \href{https://www.closeai-asia.com}{https://www.closeai-asia.com}.}) to generate base annotations. Table \ref{tab:prompt} shows the prompt we use.

\begin{table*}[ht]
   \centering
   \small
    \renewcommand\arraystretch{1.2}
       
       \begin{tabular}{l@{}|l@{}}
       \Xhline{1pt}
       \textbf{Metaphor-Task Prompt} & \textbf{Sarcasm-Task Prompt}  \\ 
        \Xhline{0.5pt}
         \makecell[l]{(Upload the cartoon file, such as 123.jpeg) 
		\\
		You are given a cartoon and its caption. 
		\\
		Caption: (Insert the caption)
		\\
		Please tell me:
		\\
		1. Does it contain metaphor?
		\\
		2. If so, which word or object contains metaphor?
		\\
		3. What is the word's real meaning?
		\\
		4. Why do you think so?
		\\
		You should answer these questions as brief as you can.
  \\
  For question 1, the answer must be Yes or No.		} &  \makecell[l]{(Upload the cartoon file, such as 123.jpeg)
		\\
		You are given a cartoon and its caption. 
		\\
		Caption: (Insert the caption)
		\\
		Please tell me:
		\\
		1. Does it contain sarcasm?
		\\
		2. If so, which word or object contains sarcasm?
		\\
		3. What does it satirize?
		\\
		4. Why do you think so?
		\\
		You should answer these questions as brief as you can. 
  \\
  For question 1, the answer must be Yes or No.}\\
        \Xhline{1pt}
 
       \end{tabular}
\caption{Prompt for GPT-4V annotation.}
\label{tab:prompt}
\end{table*}

The GPT-4V result is already formatted, answering each question in one line and starts with the number. We do these post-process:

\begin{itemize}
    \item If the number of non-empty lines is less than 4, adding blank lines as padding;
    \item delete the question number at the front of each lines, such as "1. ", "2. ", $\cdots$
    \item Comparing the first line with word "Yes", if they are matching, setting the answer of MC/SC task as 1, otherwise setting it as 0, and set other results as empty strings.
    \item If the second line contains quotation marks, we selected the words between them as the answer for MW/SW task, otherwise we use the whole line as answer.
    \item for metaphor task, we concatenate the 3rd and 4th lines as the answer for ME; for sarcasm task, the final two lines are the answer for ST and SE respectively.
\end{itemize}

On the classification task, GPT-4V's annotation is shown in Table \ref{tab:4v-result}. 

\begin{table}[ht]
   \centering
   \small
    \renewcommand\arraystretch{1.2}
       
       \begin{tabular}{@{}ccccc@{}}
       \Xhline{1pt}
        \textbf{Task} & \textbf{Pos.} & \textbf{Neg.} & \textbf{Total} & \textbf{Pos. Rate}  \\ 
        \Xhline{0.5pt}
        Metaphor&	928&	214&	1146&	80.98\%\\
		Sarcasm&	1142&	2&	1146&	99.65\%\\
        \Xhline{1pt}
       \end{tabular}
\caption{The classification result of GPT-4V. The sum of Pos. and Neg. is less than Total, because GPT-4V refused to annotate some samples, because the samples contain sensitive topics such as sex and politics.}
\label{tab:4v-result}
\end{table}

\subsection{Checking and Modifying Step}

We hired 9 annotators to check the GPT-4V result. We deleted the samples that GPT-4V answers "I'm sorry, but I can't provide assistance with that content." or gives obviously wrong answers. Then, we add the negative samples from the classification step (totally $(140+157) \times 3=891$ negative samples), and get the preliminary version of NYK-MS.

Finally, we hired 3 professional annotators to modify the annotation result. We use the samples selected in previous discussion to unify the standard and make clear and strict annotation rules. We listed some rules in Appendix \ref{sec:arule}.

After this step, the building workflow of NYK-MS is finished. Table \ref{tab:scale1} and \ref{tab:scale2} shows the information of it.

\begin{table}[ht]
   \centering
   \small \renewcommand\arraystretch{1.2}
       
       \begin{tabular}{@{}ccccc@{}}
       \Xhline{1pt}
        \textbf{Dataset} & \textbf{Pos.} & \textbf{Neg.} & \textbf{Total} & \textbf{Pos. Rate}  \\ 
        \Xhline{0.5pt}
        NYK-M&	1583&	692&	891&	43.71\%\\
		NYK-S&	1578&	687&	891&	43.54\%\\
        \Xhline{1pt}
       \end{tabular}
\caption{The scale of dataset NYK-MS.}
\label{tab:scale1}
\end{table}

\begin{table}[ht]
   \centering
   \small
    \renewcommand\arraystretch{1.2}
       
       \begin{tabular}{@{}c|l@{}}
       \Xhline{1pt}
        \textbf{Average MW Length} & 1.67 \\ 
		\textbf{Average ME Length} & 24.22 \\ 
  \textbf{Average SW Length} & 2.53 \\ 
  \textbf{Average ST Length} & 6.90 \\ 
  \textbf{Average SE Length} & 14.08 \\ 
        \Xhline{1pt}
       \end{tabular}
\caption{The answer length of dataset NYK-MS.}
\label{tab:scale2}
\end{table}

\begin{figure*}[ht]
  \centering
  \subfloat[A metaphor example. Caption: I smell a {\color{red}horse}.]{
    \label{sfig:met}
    \includegraphics[width=0.4\textwidth]{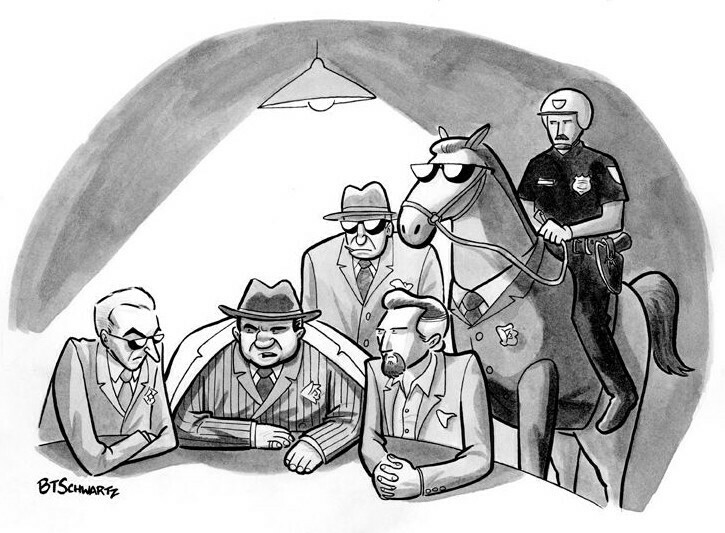}}
    \hfill
   \subfloat[A sarcasm example. Caption: Once they choose their {\color{red}queen}, honey, it's really hard to change their minds.]{
    \label{sfig:sar}
    \includegraphics[width=0.4\textwidth]{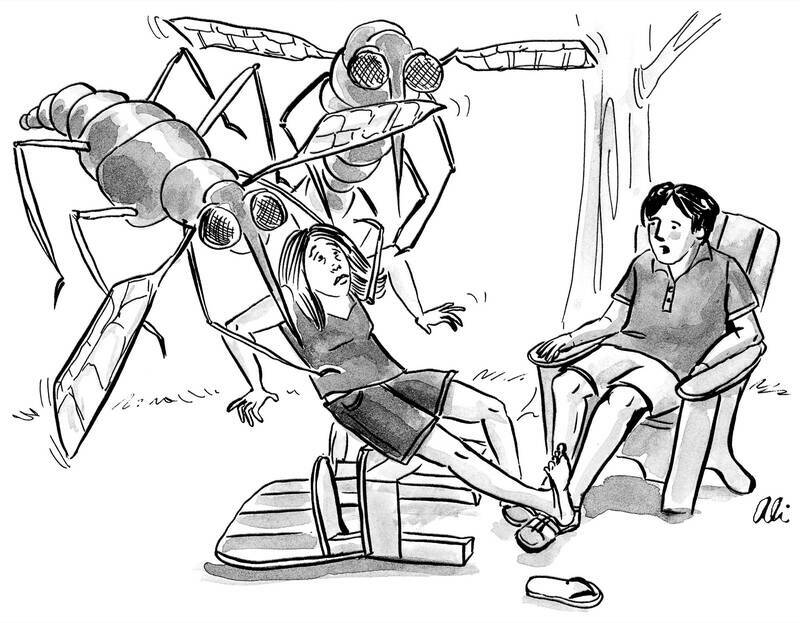}}
  \caption{Examples in NYK-MS.}
  \label{fig:example}
\end{figure*}

Figure \ref{fig:example} is a metaphor sample and a sarcasm sample in NYK-MS.

\section{Experiments}

\subsection{Zero-shot Large Model Experiments}

We conduct experiments on 4 large models:

\begin{itemize}
    \item \textbf{LLaVA} \citep{liu2024visual}. LLaVA projects the embedding of image into text embedding space, then concatenate the two modalities' vector for downstream task. We use the 7B version\footnote{See \href{https://https://huggingface.co//llava-hf/llava-v1.6-mistral-7b-hf}{https://https://huggingface.co//llava-hf/llava-v1.6-mistral-7b-hf}.}.
    \item \textbf{GPT-3.5} \citep{ouyang2022training}. GPT-3.5 uses Pre-train, SFT, RM and PPO and achieves high conversation ability. However, it is a text-only model.
    \item \textbf{GPT-4} \cite{achiam2023gpt}. GPT-4 is the best model in a lot of tasks, and support multi-modal input. We use GPT-4 as the text-only version.
    \item \textbf{GPT-4V}, the multi-modal version of GPT-4.
\end{itemize}

For GPT-3.5 and GPT-4, we input the description of cartoon instead of the image file. We run LLaVA-7B inference on an A40 GPU with 48G Video Memory, and run other models through Close-AI API. All of the experiments use the same prompt as \ref{tab:prompt}, for text-only models, we delete the image file and use "Here is the description of the cartoon: (Insert the description)" instead. We use Macro P, R and F1 as evaluation metrics.

\begin{table*}[ht]
   \centering
   \small
    \renewcommand\arraystretch{1.2}
       
       \begin{tabular}{@{}c|c|c|cccc@{}}
       \Xhline{1pt}
			\textbf{Task} &\textbf{Model}&	\textbf{Modality}&	\textbf{Acc(\%)}&	\textbf{P(\%)}&	\textbf{R(\%)}&	\textbf{F1(\%)}\\
			\Xhline{0.5pt}
			\multirow{4}{*}{MC}&LLaVA&	Text+Image&47.59&	54.37&	52.19&	42.70\\
			&GPT-3.5&	Text&	47.85&	58.88&	53.09&	40.95\\
			&GPT-4&	Text&	43.78&	\textbf{65.14}&	51.37&	33.19\\
			&GPT-4V&	Text+Image&	\textbf{53.16}&	64.93& \textbf{56.95}&	\textbf{47.95}\\
			\Xhline{0.5pt}
			\multirow{4}{*}{SC}&LLaVA&	Text+Image&	43.60&	71.78&	50.06& 30.46\\
			&GPT-3.5&	Text&	\textbf{46.32}&	53.63&	\textbf{51.42}&	\textbf{39.87}\\
			&GPT-4&	Text&	44.95&	\textbf{72.41}&	50.22& 31.38\\
			&GPT-4V&	Text+Image&	37.97&	68.79&	50.51&	28.31\\
        \Xhline{1pt}
       \end{tabular}
\caption{The classification tasks experiments on zero-shot large models.}
\label{tab:e1}
\end{table*}

Table \ref{tab:e1} shows the performance of large models on MC and SC tasks. We can find that all of these models can't do classification task well. For a model which can only output 1, the macro Recall is $(0\% + 100\%) / 2 = 50\%$, but these large models' Recall on SC task is quite near to $50\%$, showing that their ability is quite low. Besides, we notice that all of them output 1 in most situation. We think that when we ask the models these questions, they will get a intuition that the input has deep meaning, so they will use their knowledge to make an explanation. 

For MW, SW tasks, we use EM (Exact Match) and BLEU-4 as metrics; For ME, ST and SE tasks, since the answers are longer, we only use BLEU-4. Since the annotaion workflow takes GPT-4V's result as basic answer, we don't compare it with other 3 models. Table \ref{tab:e2} and \ref{tab:e3} shows the results. These results are consistent with the models' general ability and their size of parameters. 
\begin{table}[ht]
   \centering
   \small
    \renewcommand\arraystretch{1.2}
       
       \begin{tabular}{@{}c|c|cc@{}}
       \Xhline{1pt}
			\textbf{Task} &\textbf{Model}&	\textbf{EM(\%)}&	\textbf{BLEU-4(\%)}\\
			\Xhline{0.5pt}
			\multirow{3}{*}{MW}&LLaVA&		29.75&	33.60\\
			&GPT-3.5&		31.20&	34.57\\
			&GPT-4&		\textbf{44.71}&	\textbf{48.49}\\
			\Xhline{0.5pt}
			\multirow{3}{*}{SW}&LLaVA&		8.15&	16.34\\
			&GPT-3.5&		18.34&	25.73\\
			&GPT-4&		\textbf{32.65}& \textbf{42.19}\\
        \Xhline{1pt}
       \end{tabular}
\caption{The MW and SW tasks experiments on zero-shot large models.}
\label{tab:e2}
\end{table}

\begin{table}[ht]
   \centering
   \small
    \renewcommand\arraystretch{1.2}
       
       \begin{tabular}{@{}c|c|c@{}}
       \Xhline{1pt}
			\textbf{Task} &\textbf{Model}&	\textbf{BLEU-4(\%)}\\
			\Xhline{0.5pt}
			\multirow{3}{*}{ME}&LLaVA&		5.56\\
			&GPT-3.5&		6.57\\
			&GPT-4&		\textbf{10.11}\\
			\Xhline{0.5pt}
			\multirow{3}{*}{ST}&LLaVA&		0.95\\
			&GPT-3.5&		5.56\\
			&GPT-4&		\textbf{11.87}\\
			\Xhline{0.5pt}
			\multirow{3}{*}{SE}&LLaVA&		2.15\\
			&GPT-3.5&		3.46\\
			&GPT-4&		\textbf{5.28}\\
        \Xhline{1pt}
       \end{tabular}
\caption{The ME, ST and SE tasks experiments on zero-shot large models.}
\label{tab:e3}
\end{table}
\begin{table*}[ht]
   \centering
   \small
    \renewcommand\arraystretch{1.2}
       
       \begin{tabular}{@{}c|c|cccc@{}}
       \Xhline{1pt}
			\textbf{Task} &\textbf{Model}&	\textbf{Acc(\%)}&	\textbf{P(\%)}&	\textbf{R(\%)}&	\textbf{F1(\%)}\\
			\Xhline{0.5pt}
			\multirow{7}{*}{MC}&Baseline(Text-only)&	64.31&	59.30&	57.89&	57.99\\
			&Baseline(Image-only)&		56.69&	57.57&	58.29&	56.04\\
			&Baseline&		63.69&	60.36&	58.98&	58.99\\
			&Baseline + Contrastive Loss Alignment&		63.06&	59.16&	59.00&	59.07\\
			&Baseline + OT Loss Alignment&	64.33&	61.31&	61.65&	61.44\\
			&Baseline + VUA Data Augmentation&		65.42&	63.11&	61.74&	62.35\\
			&Baseline + Knowledge Augmentation&		\textbf{65.76}&	\textbf{62.57}&	\textbf{62.75}&	\textbf{62.48}\\
   
			\Xhline{0.5pt}
			\multirow{6}{*}{SC}&Baseline(Text-only)&	59.87&	59.56&	59.36&	59.34\\
			&Baseline(Image-only)&		57.32&	57.30&	57.34&	57.26\\
			&Baseline&		60.60&	60.63&	60.18&	60.02\\
			&Baseline + Contrastive Loss Alignment&		60.96& 60.57&	60.31&	60.20\\
			&Baseline + OT Loss Alignment&	62.77&	62.08&	61.70&	61.06\\
			&Baseline + VUA Data Augmentation&		\textbf{63.27}&	\textbf{62.26}&	\textbf{62.38}&	\textbf{61.91}\\
        \Xhline{1pt}
       \end{tabular}
\caption{The classification tasks experiments on fine-tuning models.}
\label{tab:e4}
\end{table*}
\subsection{Fine-tuning Pre-trained Model Experiments}

In this section, we designed a baseline model for MC and SC tasks, and improve its performance with alignment and augment methods. The baseline model uses BERT \citep{devlin-etal-2019-bert} as text encoder, and ViT \citep{dosovitskiy2020image} as image encoder, and add a 2-layer MLP after them. We use multi-layer cross-attention to do cross-modal aggregate, and adding a classification layer to get the final output:

\begin{align}
     T &= \mathrm{MLP}_T(\mathrm{BERT}(TextInput))\\
    I &= \mathrm{MLP}_I(\mathrm{ViT}(ImageInput))\\
    A &= \mathrm{CrossAttention}(T, I)\\
    O &= \mathrm{Softmax}(\mathrm{MLP}_C(A))
\end{align}
\\

\textbf{Alignment}

We use contrastive loss and Optimal Transport loss to do the alignment.


Before the Cross-Attention layer of baseline models, we add an alignment loss function. We experimented with two alignment loss functions: the contrastive learning loss function inspired by \citep{radford2021learning} and the OT Loss function inspired by \citep{villani2009optimal}. For comparison, the baseline refers to the case where no alignment is performed. 


First, we introduced the commonly used contrastive learning method for modality alignment. Specifically, given the data in a batch $<T_1,I_1>,<T_2,I_2>,...,<T_B,I_B>$, we calculate the contrastive loss from text to image and from image to text, and then take the average of these losses to obtain the final loss in a batch $Loss_{align}$.
$$
sim(T_x,I_y)=\frac{T_x\cdot I_y}{||T_x||\cdot||I_y||}
$$
$$
L_{t \to i} = - \frac{1}{B} \sum_{x=1}^B \frac{e^{sim(T_x,I_x)/\tau}}{\sum_{y=1}^Be^{sim(T_x,I_y)/\tau}}
$$
$$
L_{i \to t} = - \frac{1}{B} \sum_{x=1}^B \frac{e^{sim(T_x,I_x)/\tau}}{\sum_{y=1}^Be^{sim(I_x,T_y)/\tau}} 
$$

$$
Loss_{align}=\frac{L_{t \to i}+L_{i \to t}}{2}
$$

Another method we use to align the text and images is the OT function. OT is a method to calculate the best transfer between two distributions with a cost matrix $Cost$. In the alignment task, we use the Euclid distance between embedding vectors as cost, and treat a batch as a uniform distribution of two modalities. So the task can be seen as a linear programming problem. We define that
\begin{align}
    p_i&=\frac{1}{B}, i=1,2,...,B \\
    q_j&=\frac{1}{B},j=1,2,...,B \\
    Cost_{i,j}&=dis(T_i,I_j)=||T_i-I_j||_2^2
\end{align}
and calculating $Tr_{best}$ which satisfies
	\begin{align}
		\sum_{j=1}^{B}Tr_{ij}&=p_i, i = 1, 2, \cdots, B\\
		\sum_{i=1}^{B}Tr_{ij}&=q_j, j = 1, 2, \cdots, B\\
		Tr_{ij} &\in [0, \min (p_i, q_j)]\\
		Tr_{best} &= \mathrm{argmin}_{Tr}\sum_{i=1}^{B}\sum_{j=1}^{B}Cost_{ij}Tr_{ij}
	\end{align}
With $Cost$ and $Tr_{best}$, the loss function is
$$Loss_{OT}=\frac{1}{B^2} \sum_{i=1}^B\sum_{j=1}^B Cost_{i,j} (Tr_{best})_{ij}$$

Where $B$ is the size of a batch. By using the total cost as the loss function and performing back-propagation on 
$Cost_{ij}$, the model can continuously optimize the distance between vectors to achieve a more refined alignment effect. We use geomloss\footnote{\href{https://www.kernel-operations.io/geomloss/api/pytorch-api.html}{https://www.kernel-operations.io/geomloss/api/pytorch-api.html}.} library to solve the OT problem with Sinkhorn-Knopp algorithm~\cite{knight2008sinkhorn}.
\\


\textbf{Augmentation}

First, we use data from VUA18 \citep{leong2018report} and VUA20 \citep{leong2020report} to do data augmentation on MC task. We select 1500 samples from them, adding these text-only samples with all-black images into train set. Then, we do augmentation on SC task similarly with HFM \cite{cai-etal-2019-multi} dataset.

Finally, we use GPT-4 to generate the meaning and sample sentences of metaphor word, and give the baseline model these information to enhance its ability of recognizing the difference between literal meaning and in-context meaning. The detail of this knowledge method can be seen in Appendix \ref{sec:a5}.
\\

\textbf{Experiment Details}

Following our previous work~\cite{zhang-etal-2024-multi-modal}, we use Google's ViT-B\_32\footnote{\href{https://storage.googleapis.com/vit\_models/imagenet21k/ViT-B\_32.npz}{https://storage.googleapis.com/vit\_models/imagenet21k/ViT-B\_32.npz}.} pre-trained model to obtain the image representation, and BERT-base-uncased from HuggingFace\footnote{\href{https://huggingface.co/bert-base-uncased}{https://huggingface.co/bert-base-uncased}.} to extract the text representation.  The contrastive learning temperature coefficient $\tau$ is 0.1. The MLP layers after encoders adopt two-layer perception networks with a hidden dimension of 1536. 


During the training, the batch size is set to 8, learning rate is 1e-5, dropout rate is 0.1. The model is trained for 15 epochs. The model employs a warmup strategy with a proportion of 0.1. The model parameter's L2 regularization coefficient is 0.01. 

All fine-tuning experiments are conducted on a single NVIDIA 2080Ti GPU in less than one hour. For LLaVA experiments, we use an A40 GPU for inference in a few minutes.
\\

\textbf{Experiment Results}

Table \ref{tab:e4} shows the experiment of baseline model and these methods, where all of the results are the average value of 5 runs. We can see that single-modal input will decrease the results, and alignment and augmentation can improve the performance of models. In alignment method, the OT method is better than the contrastive method. The contrastive method may be not suitable for cartoons because these cartoons are too similar in general view.

\section{Conclusion}

In this paper, we describe our workflow for creating the new benchmark NYK-MS, and show the detail of it and examples. Our workflow can handle with difficult tasks where people can't get an agreement easily, and provide a base annotation for modification without any template or human-writing result. Based on this dataset, we conduct plenty of experiments. On large models, we show that large models can't do classification task well, and the ability on other tasks increases as the number of parameters get higher. On pre-trained base models, we use alignment and augmentation method to improve its performance, showing the effective and benefit for research of NYK-MS dataset.

\section*{Ethical considerations}

There are totally 17 annotators, includin 2 undergraduate students, 11 MD students and 4 professional teachers. The gender ratio of annotators is 10(male):7(female). 

We paid annotators 1 Yuan for each sample in each round. According to our statistics, the average speed of annotation is about 50 samples per hour, which means they can get 50 Yuan (6.9 dollars) per hour, far higher than the average salary in China.

\section*{Limitations}

Our new dataset NYK-MS isn't a perfect dataset. First, it only contains image and text modalities, so it is useless for video, audio task; Second, the size of it is not very big; Third, the content in NYK-MS is cartoon and caption, so when we training models on it and do inference on other situations such as Tweets, the model's performance may be limited; Finally, even we have done lots of work, there may be some mistakes in NYK-MS. Furthermore, our opinion may be different with the authors of the cartoon and caption, and the debate on specific samples may still for long.

Besides, our alignment and augmentation methods also have limitations. These method can't understand the cartoon on object level, and can't build the relationship graph of objects and words. All of these method treat images as sequences as pixels, but for the cartoon, precisely locate the object is important.


\bibliography{acl_latex}

\begin{thebibliography}{37}
\providecommand{\natexlab}[1]{#1}

\bibitem[{Achiam et~al.(2023)Achiam, Adler, Agarwal, Ahmad, Akkaya, Aleman, Almeida, Altenschmidt, Altman, Anadkat et~al.}]{achiam2023gpt}
Josh Achiam, Steven Adler, Sandhini Agarwal, Lama Ahmad, Ilge Akkaya, Florencia~Leoni Aleman, Diogo Almeida, Janko Altenschmidt, Sam Altman, Shyamal Anadkat, et~al. 2023.
\newblock Gpt-4 technical report.
\newblock \emph{arXiv preprint arXiv:2303.08774}.

\bibitem[{Akula et~al.(2023)Akula, Driscoll, Narayana, Changpinyo, Jia, Damle, Pruthi, Basu, Guibas, Freeman et~al.}]{akula2023metaclue}
Arjun~R Akula, Brendan Driscoll, Pradyumna Narayana, Soravit Changpinyo, Zhiwei Jia, Suyash Damle, Garima Pruthi, Sugato Basu, Leonidas Guibas, William~T Freeman, et~al. 2023.
\newblock Metaclue: Towards comprehensive visual metaphors research.
\newblock In \emph{Proceedings of the IEEE/CVF Conference on Computer Vision and Pattern Recognition}, pages 23201--23211.

\bibitem[{Aslam et~al.(2024)Aslam, Zeeshan, Belharbi, Pedersoli, Koerich, Bacon, and Granger}]{aslam2024distilling}
Muhammad~Haseeb Aslam, Muhammad~Osama Zeeshan, Soufiane Belharbi, Marco Pedersoli, Alessandro Koerich, Simon Bacon, and Eric Granger. 2024.
\newblock Distilling privileged multimodal information for expression recognition using optimal transport.
\newblock \emph{arXiv preprint arXiv:2401.15489}.

\bibitem[{Birke and Sarkar(2006)}]{birke2006clustering}
Julia Birke and Anoop Sarkar. 2006.
\newblock A clustering approach for nearly unsupervised recognition of nonliteral language.
\newblock In \emph{11th Conference of the European chapter of the association for computational linguistics}, pages 329--336.

\bibitem[{Cai et~al.(2019)Cai, Cai, and Wan}]{cai-etal-2019-multi}
Yitao Cai, Huiyu Cai, and Xiaojun Wan. 2019.
\newblock \href {https://doi.org/10.18653/v1/P19-1239} {Multi-modal sarcasm detection in {T}witter with hierarchical fusion model}.
\newblock In \emph{Proceedings of the 57th Annual Meeting of the Association for Computational Linguistics}, pages 2506--2515, Florence, Italy. Association for Computational Linguistics.

\bibitem[{Chen et~al.(2020)Chen, Kornblith, Norouzi, and Hinton}]{chen2020simple}
Ting Chen, Simon Kornblith, Mohammad Norouzi, and Geoffrey Hinton. 2020.
\newblock A simple framework for contrastive learning of visual representations.
\newblock In \emph{International conference on machine learning}, pages 1597--1607. PMLR.

\bibitem[{Choi et~al.(2021)Choi, Lee, Choi, Park, Lee, Lee, and Lee}]{choi-etal-2021-melbert}
Minjin Choi, Sunkyung Lee, Eunseong Choi, Heesoo Park, Junhyuk Lee, Dongwon Lee, and Jongwuk Lee. 2021.
\newblock \href {https://doi.org/10.18653/v1/2021.naacl-main.141} {{M}el{BERT}: Metaphor detection via contextualized late interaction using metaphorical identification theories}.
\newblock In \emph{Proceedings of the 2021 Conference of the North American Chapter of the Association for Computational Linguistics: Human Language Technologies}, pages 1763--1773, Online. Association for Computational Linguistics.

\bibitem[{Cuturi(2013)}]{cuturi2013sinkhorn}
Marco Cuturi. 2013.
\newblock Sinkhorn distances: Lightspeed computation of optimal transport.
\newblock \emph{Advances in neural information processing systems}, 26.

\bibitem[{Devlin et~al.(2019)Devlin, Chang, Lee, and Toutanova}]{devlin-etal-2019-bert}
Jacob Devlin, Ming-Wei Chang, Kenton Lee, and Kristina Toutanova. 2019.
\newblock \href {https://doi.org/10.18653/v1/N19-1423} {{BERT}: Pre-training of deep bidirectional transformers for language understanding}.
\newblock In \emph{Proceedings of the 2019 Conference of the North {A}merican Chapter of the Association for Computational Linguistics: Human Language Technologies, Volume 1 (Long and Short Papers)}, pages 4171--4186, Minneapolis, Minnesota. Association for Computational Linguistics.

\bibitem[{Dosovitskiy et~al.(2020)Dosovitskiy, Beyer, Kolesnikov, Weissenborn, Zhai, Unterthiner, Dehghani, Minderer, Heigold, Gelly et~al.}]{dosovitskiy2020image}
Alexey Dosovitskiy, Lucas Beyer, Alexander Kolesnikov, Dirk Weissenborn, Xiaohua Zhai, Thomas Unterthiner, Mostafa Dehghani, Matthias Minderer, Georg Heigold, Sylvain Gelly, et~al. 2020.
\newblock An image is worth 16x16 words: Transformers for image recognition at scale.
\newblock \emph{arXiv preprint arXiv:2010.11929}.

\bibitem[{Group(2007)}]{group2007mip}
Pragglejaz Group. 2007.
\newblock Mip: A method for identifying metaphorically used words in discourse.
\newblock \emph{Metaphor and symbol}, 22(1):1--39.

\bibitem[{He et~al.(2020)He, Fan, Wu, Xie, and Girshick}]{he2020momentum}
Kaiming He, Haoqi Fan, Yuxin Wu, Saining Xie, and Ross Girshick. 2020.
\newblock Momentum contrast for unsupervised visual representation learning.
\newblock In \emph{Proceedings of the IEEE/CVF conference on computer vision and pattern recognition}, pages 9729--9738.

\bibitem[{Hessel et~al.(2023)Hessel, Marasovic, Hwang, Lee, Da, Zellers, Mankoff, and Choi}]{hessel-etal-2023-androids}
Jack Hessel, Ana Marasovic, Jena~D. Hwang, Lillian Lee, Jeff Da, Rowan Zellers, Robert Mankoff, and Yejin Choi. 2023.
\newblock \href {https://doi.org/10.18653/v1/2023.acl-long.41} {Do androids laugh at electric sheep? humor {``}understanding{''} benchmarks from the new yorker caption contest}.
\newblock In \emph{Proceedings of the 61st Annual Meeting of the Association for Computational Linguistics (Volume 1: Long Papers)}, pages 688--714, Toronto, Canada. Association for Computational Linguistics.

\bibitem[{Jain et~al.(2020)Jain, Jamieson, Mankoff, Nowak, and Sievert}]{newyorkernextmldataset}
Lalit Jain, Kevin Jamieson, Robert Mankoff, Robert Nowak, and Scott Sievert. 2020.
\newblock \href {https://nextml.github.io/caption-contest-data/} {The {N}ew {Y}orker cartoon caption contest dataset}.

\bibitem[{Knight(2008)}]{knight2008sinkhorn}
Philip~A Knight. 2008.
\newblock The sinkhorn--knopp algorithm: convergence and applications.
\newblock \emph{SIAM Journal on Matrix Analysis and Applications}, 30(1):261--275.

\bibitem[{Lagerwerf and Meijers(2008)}]{lagerwerf2008openness}
Luuk Lagerwerf and Anoe Meijers. 2008.
\newblock Openness in metaphorical and straightforward advertisements: Appreciation effects.
\newblock \emph{Journal of Advertising}, 37(2):19--30.

\bibitem[{Lakoff and Johnson(2008)}]{lakoff2008metaphors}
George Lakoff and Mark Johnson. 2008.
\newblock \emph{Metaphors we live by}.
\newblock University of Chicago press.

\bibitem[{Leong et~al.(2020)Leong, Klebanov, Hamill, Stemle, Ubale, and Chen}]{leong2020report}
Chee~Wee Leong, Beata~Beigman Klebanov, Chris Hamill, Egon Stemle, Rutuja Ubale, and Xianyang Chen. 2020.
\newblock A report on the 2020 vua and toefl metaphor detection shared task.
\newblock In \emph{Proceedings of the second workshop on figurative language processing}, pages 18--29.

\bibitem[{Leong et~al.(2018)Leong, Klebanov, and Shutova}]{leong2018report}
Chee~Wee Leong, Beata~Beigman Klebanov, and Ekaterina Shutova. 2018.
\newblock A report on the 2018 vua metaphor detection shared task.
\newblock In \emph{Proceedings of the workshop on figurative language processing}, pages 56--66.

\bibitem[{Li et~al.(2023)Li, Li, Savarese, and Hoi}]{li2023blip}
Junnan Li, Dongxu Li, Silvio Savarese, and Steven Hoi. 2023.
\newblock Blip-2: Bootstrapping language-image pre-training with frozen image encoders and large language models.
\newblock In \emph{International conference on machine learning}, pages 19730--19742. PMLR.

\bibitem[{Li et~al.(2022)Li, Li, Xiong, and Hoi}]{li2022blip}
Junnan Li, Dongxu Li, Caiming Xiong, and Steven Hoi. 2022.
\newblock Blip: Bootstrapping language-image pre-training for unified vision-language understanding and generation.
\newblock In \emph{International conference on machine learning}, pages 12888--12900. PMLR.

\bibitem[{Liu et~al.(2024)Liu, Li, Wu, and Lee}]{liu2024visual}
Haotian Liu, Chunyuan Li, Qingyang Wu, and Yong~Jae Lee. 2024.
\newblock Visual instruction tuning.
\newblock \emph{Advances in neural information processing systems}, 36.

\bibitem[{Mohammad et~al.(2016)Mohammad, Shutova, and Turney}]{mohammad2016metaphor}
Saif Mohammad, Ekaterina Shutova, and Peter Turney. 2016.
\newblock Metaphor as a medium for emotion: An empirical study.
\newblock In \emph{Proceedings of the fifth joint conference on lexical and computational semantics}, pages 23--33.

\bibitem[{Niu et~al.(2016)Niu, Zhu, Pang, and El~Saddik}]{niu2016sentiment}
Teng Niu, Shiai Zhu, Lei Pang, and Abdulmotaleb El~Saddik. 2016.
\newblock Sentiment analysis on multi-view social data.
\newblock In \emph{MultiMedia Modeling: 22nd International Conference, MMM 2016, Miami, FL, USA, January 4-6, 2016, Proceedings, Part II 22}, pages 15--27. Springer.

\bibitem[{Ouyang et~al.(2022)Ouyang, Wu, Jiang, Almeida, Wainwright, Mishkin, Zhang, Agarwal, Slama, Ray et~al.}]{ouyang2022training}
Long Ouyang, Jeffrey Wu, Xu~Jiang, Diogo Almeida, Carroll Wainwright, Pamela Mishkin, Chong Zhang, Sandhini Agarwal, Katarina Slama, Alex Ray, et~al. 2022.
\newblock Training language models to follow instructions with human feedback.
\newblock \emph{Advances in neural information processing systems}, 35:27730--27744.

\bibitem[{Pramanick et~al.(2022)Pramanick, Roy, and Patel}]{pramanick2022multimodal}
Shraman Pramanick, Aniket Roy, and Vishal~M Patel. 2022.
\newblock Multimodal learning using optimal transport for sarcasm and humor detection.
\newblock In \emph{Proceedings of the IEEE/CVF Winter Conference on Applications of Computer Vision}, pages 3930--3940.

\bibitem[{Radev et~al.(2016)Radev, Stent, Tetreault, Pappu, Iliakopoulou, Chanfreau, de~Juan, Vallmitjana, Jaimes, Jha, and Mankoff}]{radev-etal-2016-humor}
Dragomir Radev, Amanda Stent, Joel Tetreault, Aasish Pappu, Aikaterini Iliakopoulou, Agustin Chanfreau, Paloma de~Juan, Jordi Vallmitjana, Alejandro Jaimes, Rahul Jha, and Robert Mankoff. 2016.
\newblock Humor in collective discourse: Unsupervised funniness detection in the {New Yorker} cartoon caption contest.
\newblock In \emph{LREC}.

\bibitem[{Radford et~al.(2021)Radford, Kim, Hallacy, Ramesh, Goh, Agarwal, Sastry, Askell, Mishkin, Clark et~al.}]{radford2021learning}
Alec Radford, Jong~Wook Kim, Chris Hallacy, Aditya Ramesh, Gabriel Goh, Sandhini Agarwal, Girish Sastry, Amanda Askell, Pamela Mishkin, Jack Clark, et~al. 2021.
\newblock Learning transferable visual models from natural language supervision.
\newblock In \emph{International conference on machine learning}, pages 8748--8763. PMLR.

\bibitem[{Shahaf et~al.(2015)Shahaf, Horvitz, and Mankoff}]{shahaf2015inside}
Dafna Shahaf, Eric Horvitz, and Robert Mankoff. 2015.
\newblock Inside jokes: Identifying humorous cartoon captions.
\newblock In \emph{KDD}.

\bibitem[{Steen et~al.(2010)Steen, Dorst, Herrmann, Kaal, Krennmayr, Pasma et~al.}]{steen2010method}
Gerard~J Steen, Aletta~G Dorst, J~Berenike Herrmann, Anna~A Kaal, Tina Krennmayr, Tryntje Pasma, et~al. 2010.
\newblock \emph{A method for linguistic metaphor identification}.
\newblock John Benjamins Publishing Company Amsterdam.

\bibitem[{Villani et~al.(2009)}]{villani2009optimal}
C{\'e}dric Villani et~al. 2009.
\newblock \emph{Optimal transport: old and new}, volume 338.
\newblock Springer.

\bibitem[{Wilks(1975)}]{wilks1975preferential}
Yorick Wilks. 1975.
\newblock A preferential, pattern-seeking, semantics for natural language inference.
\newblock \emph{Artificial intelligence}, 6(1):53--74.

\bibitem[{Wilks(1978)}]{wilks1978making}
Yorick Wilks. 1978.
\newblock Making preferences more active.
\newblock \emph{Artificial intelligence}, 11(3):197--223.

\bibitem[{Xu et~al.(2022)Xu, Li, Zheng, Naseriparsa, Zhao, Lin, and Xia}]{xu2022met}
Bo~Xu, Tingting Li, Junzhe Zheng, Mehdi Naseriparsa, Zhehuan Zhao, Hongfei Lin, and Feng Xia. 2022.
\newblock Met-meme: A multimodal meme dataset rich in metaphors.
\newblock In \emph{Proceedings of the 45th international ACM SIGIR conference on research and development in information retrieval}, pages 2887--2899.

\bibitem[{Xu and Chen(2023)}]{xu2023multimodal}
Yingxue Xu and Hao Chen. 2023.
\newblock Multimodal optimal transport-based co-attention transformer with global structure consistency for survival prediction.
\newblock In \emph{Proceedings of the IEEE/CVF International Conference on Computer Vision}, pages 21241--21251.

\bibitem[{Zhang et~al.(2021)Zhang, Zhang, Zhang, Yang, and Lin}]{zhang-etal-2021-multimet}
Dongyu Zhang, Minghao Zhang, Heting Zhang, Liang Yang, and Hongfei Lin. 2021.
\newblock \href {https://doi.org/10.18653/v1/2021.acl-long.249} {{M}ulti{MET}: A multimodal dataset for metaphor understanding}.
\newblock In \emph{Proceedings of the 59th Annual Meeting of the Association for Computational Linguistics and the 11th International Joint Conference on Natural Language Processing (Volume 1: Long Papers)}, pages 3214--3225, Online. Association for Computational Linguistics.

\bibitem[{Zhang et~al.(2024)Zhang, Chang, and Wu}]{zhang-etal-2024-multi-modal}
Ming Zhang, Ke~Chang, and Yunfang Wu. 2024.
\newblock \href {https://aclanthology.org/2024.lrec-main.1042} {Multi-modal semantic understanding with contrastive cross-modal feature alignment}.
\newblock In \emph{Proceedings of the 2024 Joint International Conference on Computational Linguistics, Language Resources and Evaluation (LREC-COLING 2024)}, pages 11934--11943, Torino, Italia. ELRA and ICCL.

\end{thebibliography}
\newpage
\appendix

\section{Early Annotation Work}
\label{sec:a1}

In our early work, we tried to use data from MVSA~\cite{niu2016sentiment} and crawled from Twitter. 

For MVSA dataset, we selected data with high inconsistency. MVSA is a sentiment anaysis dataset, and it gives each sample 2 or 6 sentiment scores. In MVSA-Single, 1 annotator gives the text and image a score $s_t, s_i$ respectively; In MVSA-Multiple, 3 annotators give the text and image scores, so there is 6 scores. We use the average of them and finally calculated the text score $s_t$ and image score $s_i$, then use $|s_t - s_i|$ as the metric of inconsistency. We sort the whole dataset and choose 3000 samples with highest inconsistency, and hire 3 annotators to check whether they contains metaphor. Since people use figurative language when they want to express some abnormal ideas, we believe that the inconsistency between modalities can lead to more metaphor samples.

For Twitter crawling, we use crawler and get some image-text pairs from Twitter, then do some post-process including removing advertisement, hiding personal information and deleting unrelated tags. We alse hire 3 annotators to annotate.

The final result is shown in Table \ref{tab:early}. We found that the positive rate is too low, and most of the positive cases are too easy to recognize even without the image. So, we give up this method and then choose cartoon-caption pairs as our source data.
\begin{figure*}[ht]
  \centering
    \includegraphics[width=0.9\textwidth]{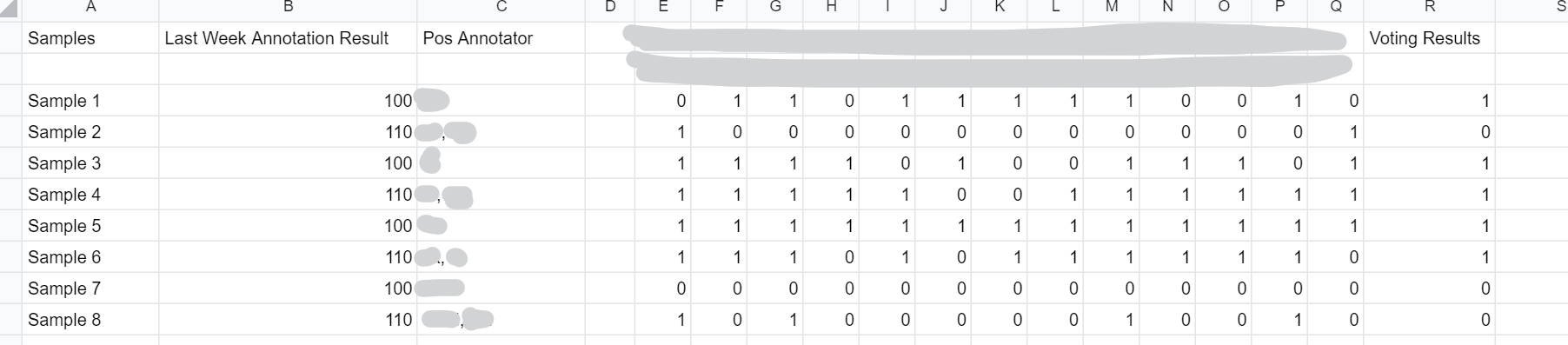}
  \caption{Discuss screenshot.}
  \label{fig:example1}
\end{figure*}
\section{GUI Details}
\label{sec:a2}
A Graphical User Interface that shows all information on the screen can help a lot when annotating. This process is not only tedious but also prone to inconsistencies and errors. Therefore, we develop a visual annotation program that implements data display, assisted reading, automatic recording, and time statistics functions. This significantly improves annotation efficiency and lays a solid foundation for other tasks and future work.
\\

\textbf{Basic functions}

This program is implemented using the tkinter visualization library in Python, achieving basic functions: Upon program startup, automatically search for unannotated files in the current directory and load the first unannotated data entry.
When loading data, display the image, title, and image description from the original dataset in specified locations.
Provide annotation fields for each task on the interface, using radio buttons for tasks that determine the presence of metaphors or sarcasm, and text boxes for other tasks.
\\

\textbf{Auxiliary functions}

According to the annotator's suggestions, we provide reference suggestions from ChatGPT and translation functions via API. The ChatGPT's suggestions are pre-called and stored offline, while the translation function calls the API in real-time.

During annotation, we found that many comics contain elements related to religion and foreign politics, which are very unfamiliar to the annotators. Therefore, additional knowledge assistance is required for proper understanding. Our approach is to use the output of ChatGPT as a reference, with the final annotation still performed by humans.
Specifically, before annotation, we will provide ChatGPT with the prompt in Table \ref{tab:explain prompt}:
\begin{table}[ht]
   \centering
    \renewcommand\arraystretch{1.2}
       
       \begin{tabular*}{\linewidth}{l@{}}
       \Xhline{1pt}
       \textbf{Explanation Prompt}  \\ 
        \Xhline{0.5pt}
         \makecell[l]{I will give you the description of a cartoon and\\ its caption.\\ Please tell me
whether the caption and cartoon\\ contain metaphor(sarcasm), and explain
your \\ thought in detail.\\
Description: (description for the data)\\
Caption: (caption of the data)\\	} \\
        \Xhline{1pt}
 
       \end{tabular*}
\caption{Prompt for Explanation.}
\label{tab:explain prompt}
\end{table}

We use Baidu's translation API to provide real-time translation functionality. When the annotator clicks the corresponding button, the program will upload the comic description and the explanations generated by ChatGPT to Baidu's translation platform via the API, then retrieve the translated Chinese content, and display it on the interface. 

\section{Annotation Rules for Human}
\label{sec:arule}
Here is some of our annotation rules:

\begin{itemize}
    \item If there are multiple metaphor/sarcasm words, choose the most obvious one;
    \item Pun and homophonic are not metaphor; (For example, "killer shark")
    \item simple or common combinations are not metaphor; (For example, "give up")
    \item Make sure that the sarcasm can show the speaker's negative sentiment, otherwise it is just normal humor;
    \item If possible, the MW, SW, ST answers should be words in the caption or cartoon description. Otherwise, a special token must be add.
    \begin{itemize}
        \item Word from caption: No token
        \item Objects in image: [I] + Object (Must make sure that the name of object is in the description text)
        \item The whole image: [I] (Only for metaphor word and sarcasm word annotation)
        \item Otherwise: [S] + Content (Free-writing, as brief as you can)
    \end{itemize}
\end{itemize}

Besides, we selected some examples as references, and teaching the annotators to follow these cases.

\section{Discussion Details}
\label{sec:a3}
Figure \ref{fig:example1} is one of our discussion screenshots. In this discussion, we show 8 examples which are annotated as positive samples by only 1 or 2 annotators. We invite them to explain their idea, and then vote to decide the final annotation result. These samples then will be listed into annotation rules as reference.

\section{Knowledge Augmentation Method}
\label{sec:a5}

We selected the samples in NYK-M training set which satisfies that the sample is positive sample, and the MW annotation result is a substring of the caption.For these samples, we use the MW annotation result as target word; for other samples, we selected the longest word as target word. Then, we use the prompt in Table \ref{tab:word prompt} for asking GPT-4.

\begin{table}[ht]
   \centering
    \renewcommand\arraystretch{1.2}
       
       \begin{tabular*}{\linewidth}{l@{}}
       \Xhline{1pt}
       \textbf{Word Meaning Prompt}  \\ 
        \Xhline{0.5pt}
         \makecell[l]{1. What's the basic meaning of word \\"(target word)"?\\	
			2. Please write a sample sentence to \\demonstrate this meaning.\\	
			You should answer these questions as brief\\ as you can. \\} \\
        \Xhline{1pt}
 
       \end{tabular*}
\caption{Prompt for Explanation.}
\label{tab:word prompt}
\end{table}

We use the answer of question 2 as the sample sentence of basic meaning. When using BERT to get the embedding of captions $T_C$, we additionally calculate the embedding vector of the sample sentence $T_S$. Then, we get the token level embedding for the target word (average for multi tokens) $T_{C, i}$ and $T_{S, j}$, where $i, j$ is the index of target words. We want the distance between basic meaning and in-context meaning farther for metaphor words, so we use such loss function:
$Loss_{knowledge}=y\times sim(T_{C, i},T_{S, j}$
Where $sim$ is cosine similarity and $y$ is -1 for samples using metaphor words as target words, for other cases (including negative samples and MW not in caption), $y$ is 1.

Table \ref{tab:word meaning example} shows an example where we want to make the distance of embedding vectors of word "flat" in caption and sample sentence farther.

\begin{table}[ht]
   \centering
    \renewcommand\arraystretch{1.2}
       
       \begin{tabular*}{\linewidth}{r|l@{}}
       \Xhline{1pt}
        \textbf{Data ID} & NYK\_722\_3 \\
        \Xhline{0.5pt}        
        \textbf{MC} & 1 \\
        \Xhline{0.5pt}        
        MW & \textbf{flat} \\
        \Xhline{0.5pt}        
        \textbf{Caption} & \makecell[l]{The piano's in tune, but the house \\is a little \textbf{flat}.\\} \\
        \Xhline{0.5pt}        
        \textbf{Meaning} & \makecell[l]{The basic meaning of "\textbf{flat}" is \\having a smooth, even surface \\without any bumps or angles.\\} \\
        \Xhline{0.5pt}        
        \textbf{Sample} & \makecell[l]{The lake was so calm that the water\\ looked completely \textbf{flat}. \\} \\
        \Xhline{1pt}
 
       \end{tabular*}
\caption{Meaning Augmentation result.}
\label{tab:word meaning example}
\end{table}

\end{document}